\def\year{2020}\relax
\documentclass[letterpaper]{article} 
\usepackage{aaai20}  
\usepackage{times}  
\usepackage{helvet} 
\usepackage{courier}  
\usepackage[hyphens]{url}  
\usepackage{graphicx} 
\usepackage{graphicx}  
\usepackage{multirow}
\usepackage{amsmath}
\usepackage{amssymb}
\usepackage{wasysym}
\usepackage{rotating}

\newcommand{\parencite}[1]{(\citeauthor{#1} \citeyear{#1})}
\providecommand{\tabularnewline}{\\}
\DeclareFontEncoding{LGR}{}{}
\DeclareRobustCommand{\greektext}{%
	\fontencoding{LGR}\selectfont\def\encodingdefault{LGR}}
\DeclareRobustCommand{\textgreek}[1]{\leavevmode{\greektext #1}}
\ProvideTextCommand{\~}{LGR}[1]{\char126#1}

\title{ElixirNet: Relation-aware Network Architecture Adaptation for Medical Lesion Detection}
\author{Chenhan Jiang,\textsuperscript{\rm 2 \thanks{Both authors contributed equally to this work.}} Shaoju Wang,\textsuperscript{\rm 1 $^{*}$} Hang Xu,\textsuperscript{\rm 2} Xiaodan Liang,\textsuperscript{\rm 1}\thanks{Corresponding Author: xdliang328@gmail.com}\\ \Large \textbf{Nong Xiao\textsuperscript{\rm 1} } \\ \textsuperscript{\rm 1}Sun Yat-Sen University \textsuperscript{\rm 2}Huawei Noah's Ark Lab}

\begin{document}
\maketitle
\begin{abstract}
Most advances in medical lesion detection network are limited to subtle
modification on the conventional detection network designed for natural
images. However, there exists a vast domain gap between medical images
and natural images where the medical image detection often suffers
from several domain-specific challenges, such as high lesion/background
similarity, dominant tiny lesions, and severe class imbalance. Is
a hand-crafted detection network tailored for natural image undoubtedly
good enough over a discrepant medical lesion domain? Is there more
powerful operations, filters, and sub-networks that better fit the
medical lesion detection problem to be discovered? In this paper,
we introduce a novel ElixirNet that includes three components: 1)
TruncatedRPN balances positive and negative data for false positive
reduction; 2) Auto-lesion Block is automatically customized for medical
images to incorporates relation-aware operations among region proposals,
and leads to more suitable and efficient classification and localization.
3) Relation transfer module incorporates the semantic relationship
and transfers the relevant contextual information with an interpretable
graph, thus alleviates the problem of lack of annotations for all
types of lesions. Experiments on DeepLesion and Kits19 prove the effectiveness
of ElixirNet, achieving improvement of both sensitivity and precision
over FPN with fewer parameters.
\end{abstract}

\section{Introduction}

\begin{figure}[t]
\begin{centering}
\includegraphics[scale=0.28]{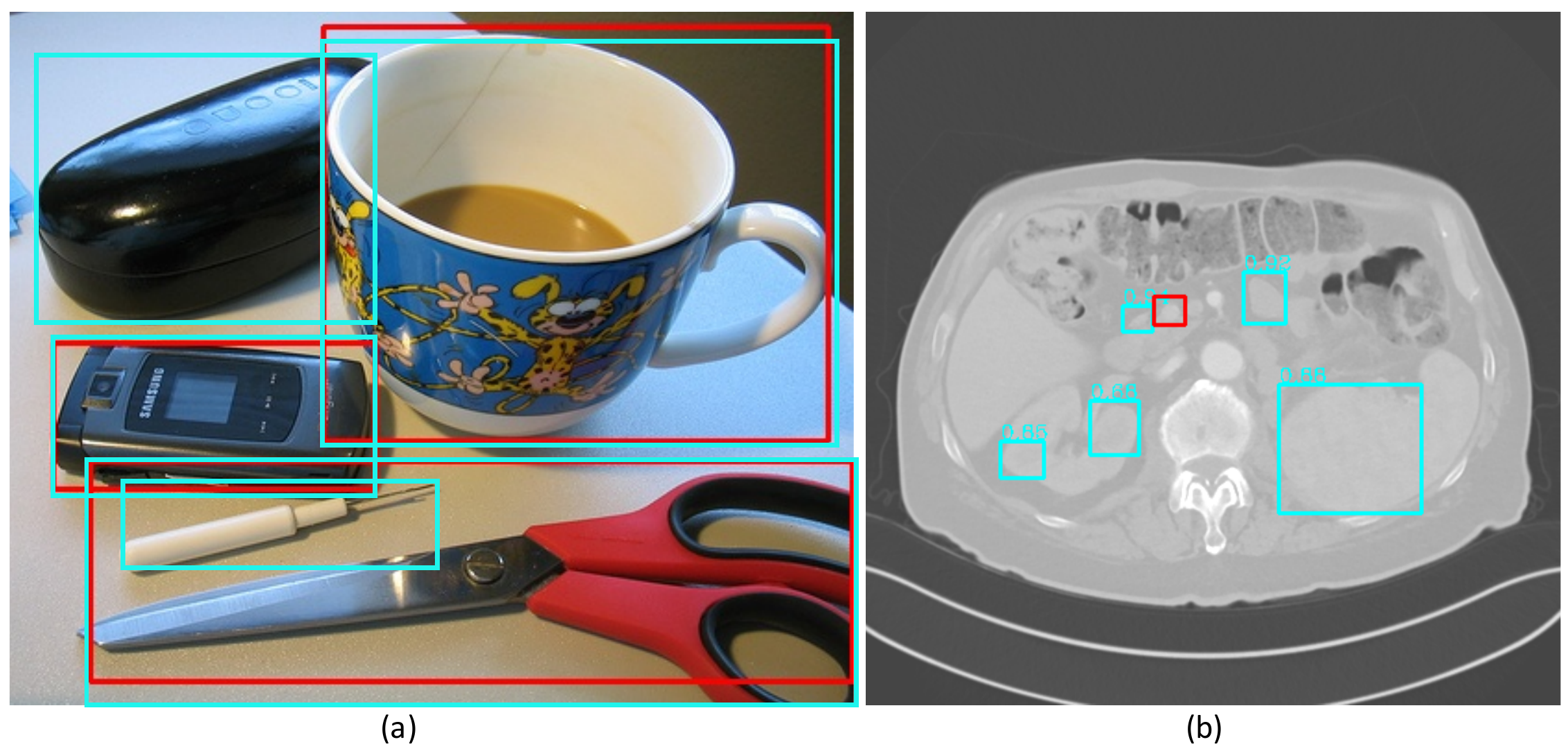}
\par\end{centering}
\caption{\label{fig:introduction-1}A vast domain gap between medical and natural
images. FPN works well on MSCOCO (a), while it fails on DeepLesion
(b) with the tiny-size lesion, identical nearby regions and overmuch
false positive (ground-truth is red box).}
\end{figure}

Lesion detection in medical CT images is an important prerequisite
for computer-aided detection/diagnosis (CADe/CADx). Recently, remarkable
progress has been brought to the application of deep learning paradigms,
especially Convolutional Neural Network (CNN) \cite{krizhevsky2012imagenet},
to CADe/CADx \cite{yan20183d,wang2017zoom,greenspan2016guest}. Most
of the works in this area directly use the natural image detection
pipelines without adapting to the medical imaging domain. \cite{shin2016deep}
transfers the CNN model with pre-trained ImageNet to medical images
detection directly. \cite{jaeger2018retina} fuses RetinaNet detector
\parencite{lin2017focal} with the U-Net \parencite{ronneberger2015u}
architecture to improve detection performance with full segmentation
supervision. Moreover, \cite{yan20183d} sends multiple neighbouring
slices into the R-FCN \cite{dai2016r} to generate feature maps separately,
which are then aggregated for final prediction. Those methods finetune
common detection framework on medical datasets without designing a
special network for medical images. Different from natural images,
only one to three lesions exist in a medical image and look similar
to the nearby non-lesions. Also, lesions are usually tiny-sized and
severe class imbalanced, illustrated in Figure \ref{fig:introduction-1}.
Using the conventional detection network for natural images directly
is inefficient and will lead to a performance drop in those scenarios.
Thus, customizing networks for medical lesion detection
is in great need.

\begin{figure*}[t]
\begin{centering}
\includegraphics[scale=0.35]{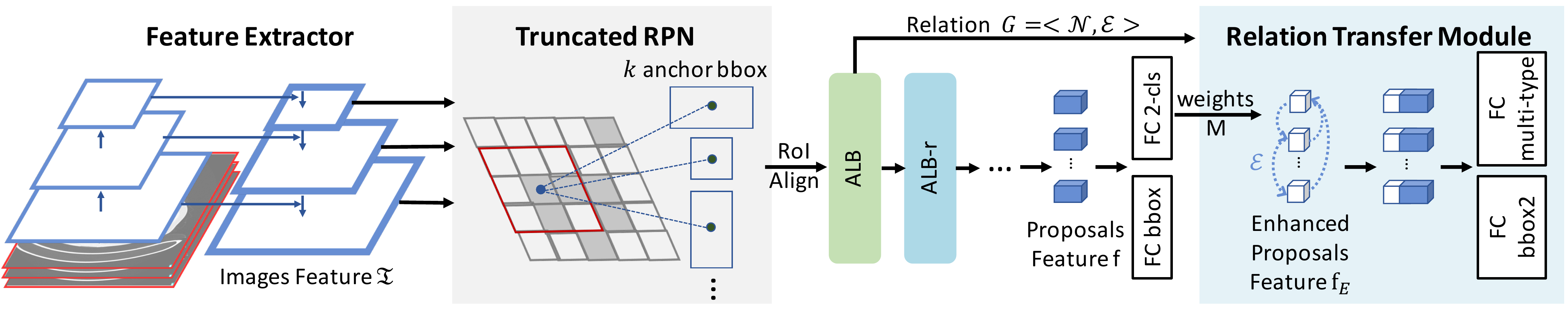}
\par\end{centering}
\caption{\label{fig:framework}An overview of ElixirNet. TruncatedRPN is introduced
to avoid overmuch false. Auto-lesion Blocks (ALB and ALB$_{r}$) are
found by differentiable NAS to capture proposals representation and
relation $\mathcal{E}$. Relation transfer module encodes more contextual
information and transfers lesion presence embeddings $M$ to enhanced
proposal feature by learned relation $\mathcal{E}$.}
\end{figure*}

Conventional natural images detection pipelines most consist of three
components: feature extractor; region proposal network (RPN) and RCNN
head for processing proposals feature. While it is well-known that
ImageNet pretrained feature extractor \parencite{russakovsky2015imagenet}
is beneficial to medical images networks \parencite{tajbakhsh2016convolutional,shin2016deep},
RPN and RCNN head designed for natural images are inconsistent for
medical images. To maintain a sufficiently high recall for proposals
in natural images, uniformly sampling anchors over the whole localization
in RPN leads to severe false candidates in medical images. \cite{roth2015improving}
trained classifiers on the aggregation of multiple 2D slices for false
positive reduction, but it has two stages and is not end-to-end. We
introduce TruncatedRPN (TRPN) to adaptively localize the lesion object
for different input. RCNN heads in \cite{lin2017feature,ren2015faster}
are too simple to exact discriminate lesion from the similar nearby
non-lesion region. \cite{liu2018receptive} considers various receptive
fields to mimic human visual systems but its neglect of relational
interaction. Relation Network \parencite{hu2018relation} designed
an adapted attention head to incorporate relationships from relevant
proposals of which performance is limited by a single receptive field.

Recently neural architecture network search (NAS) has achieved much
success in developing architectures that outperform hand-craft architectures
on image classification and semantic segmentation task \parencite{nekrasov2018fast,de2018automated,zoph2018learning}.
Different from \cite{zoph2018learning} on natural object detection
that only transfers searched classification architecture as feature
extractor and suffers from the large cost of GPU hours, we propose
a novel search space for medical images following differentiable NAS
\parencite{liu2018darts}, which optimizes architecture parameters
based on the gradient descent. In this paper, we automatically customize
an Auto-lesion block (ALB) as RCNN head substitute, which includes
flexible receptive fields to locate the tiny-size lesion, relation
context to identify similar regions and enough complexity to capture
good feature representation.

Current lesion detection methods generally target one particular disease
or lesion. Yet multiple types of lesions may appear on the same CT
slice, e.g. abdomen/liver/kidney are neighboring types. Multiple correlated
findings can help for better prediction and are crucial to building
an automatic radiological diagnosis and reasoning system. However,
it remains a challenge to develop a universal multi-purpose CAD framework
due to rare annotations for all types of lesions, capable of detecting
multiple disease types seamlessly. Compared with finetuning a new
multi-type classifier, we introduce Relation Transfer Module (RTM)
to incorporate the semantic relation learned by ALB and transfers
the relevant contextual information with an interpretable graph in
the end-to-end training. In a word, our ElixirNet consisted of TruncatedRPN,
Auto-lesion block and relation transfer module can alleviate problems
on medical lesion domain and enable a unified lesion detection network
seamlessly.

In our experiments on DeepLesion \parencite{yan2018deeplesion}, we
observe consistent gains on the 3 different common feature extractor
with reducing almost one-third of the parameters. The search is very
efficient even for $512\times512$ input and only takes about 2 days
on 4 GTX1080. The sensitivity of lesion detection with 3 false positives
per image has about 2.3\% improvement based on FPN. And for multi-type
lesion detection, consistent improvement of sensitivity on all kinds
of lesion can be found.

\section{Related Work}

\begin{figure*}
\begin{centering}
\includegraphics[scale=0.6]{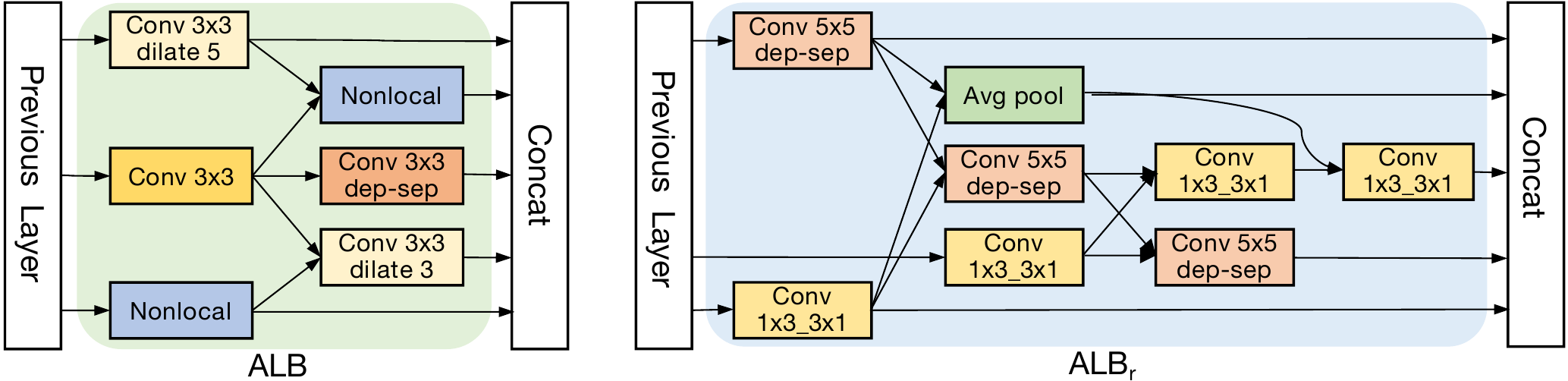}
\par\end{centering}
\caption{\label{fig:cell}The detailed architecture of ALB and ALB$_{r}$.
ALB keeps the same spatial resolution and output channels with input
including flexible dilated convolution to locate tiny-size lesion
and Non-local operator to incorporate relation-aware information among
lesions. To deepen network, an inception-like block ALB$_{r}$ is
designed for double channels and resolution reduction.}
\end{figure*}

\textbf{Automated Lesion Detection. }The detection of lesions in images
is a key part of diagnosis and is one of the most labor-intensive
for clinicians. Typically, the tasks consist of the localization and
identification of small lesions in the full image space. Different
from common object detection, lesion detection needs to consider 3D
context information while facing more challenging problems about serve
imbalance of categories and tiny-size objects. Recently, great progress
has been brought to medical imaging analysis by deep learning methods
\parencite{liao2019evaluate,yan2018deep,yu2018recurrent}. Many conventional
methods which are based on hand-crafted local features have been replaced
by deep neural networks, which typically produces higher detection
accuracy \cite{greenspan2016guest}. However, most lesion detection
\parencite{jaeger2018retina,wang2017zoom,yan20183d,yan2018deeplesion}
enhanced by CNN directly extend existing detection frameworks, which
are typically designed for natural images, to process volumetric CT
data. \cite{liao2019evaluate} uses small 3D patches of CT data as
the input of the region proposal network (RPN) \cite{ren2015faster}.
Another method \cite{yan20183d} sends multiple neighboring slices
into R-FCN \cite{dai2016r} to generate feature maps separately, which
are then aggregated for final prediction. These methods do not focus
on designing specific network for medical lesion detection.

\textbf{Neural Architecture Search Method. }Neural Architecture Search
(NAS) aims at automatically designing finding the optimal architecture
in a given search space such that the validation accuracy is maximized
on the given task. Conventional architecture search methods can be
categorized as random with weights prediction \cite{brock2017smash},
evolution \cite{miikkulainen2019evolving}, and reinforcement learning
\cite{zoph2016neural,zoph2018learning} , which have achieved highly
competitive performance on image classification . Only few of them
focus on object detection. DetNAS \cite{chen2019detnas} use the one-shot
NAS to design a new feature extractor, while ImageNet pretrained feature
extractor is beneficial to medical image networks. NAS-FPN \cite{ghiasi2019fpn}
only focuses on feature fusion connection in the feature extractor
and uses RL-based search algorithm. The search is very time-consuming
even though a proxy task with ResNet-10 backbone. More efficient solutions
for NAS have been recently proposed. Instead of searching over a discrete
set of candidate architectures, \cite{liu2018darts,liu2019auto} relax
the search space to be continuous, so that the architecture can be
optimized concerning its validation set performance by gradient descent.
Our work follows the differentiable NAS formulation and extends it
into the specific medical lesion detection setting.

\section{The Proposed Approach}

Our ElixirNet is based on two stage detection pipeline with feature
pyramid extractor. Since lesion detection relies on 3D context information
and current detectors are mostly designed with a three channels input
(RGB), we naturally group the neighboring three slices to 3-channel
images and thus implement current detection backbone with pretrained
weights on ImageNet \cite{russakovsky2015imagenet}. As shown in Figure
\ref{fig:framework}, we introduce TruncatedRPN to avoid overmuch
false positive prediction by restraining some regions where lesions
are unlikely to exist. Then the output of RoIAlign is feed into Auto-lesion
block (ALB) , which is found by differentiable NAS according to a
novel search space based on observation of many popular designs and
medical images characteristics. After aggregating information from
various receptive fields and learning a region-to-region undirected
graph $\boldsymbol{G}$ : $\boldsymbol{\boldsymbol{G}}=<\mathcal{N},\mathcal{\mathcal{E}}>$
in ALB, we further propose a relation transfer module to encode more
contextual information and transfer coarse embeddings of lesion exsiting
to enhance proposal feature by learned relation $\mathcal{E}$. In
what follows, we describe the structure of the above mentioned three
modules and explain their implementation in detail.

\subsection{TruncatedRPN}

RPN is basis in two-stage nature image object detection pipelines,
which takes image feature from feature extractor and outputs a set
of rectangular object proposals with objectness score by sliding-window
at each location. And it supposes a uniform distribution of anchor
localization for all input. This method is inefficient for medical
images, as rare lesion objects exist on the image and many of anchors
are placed in regions where lesion objects are impossible to arise.
In this work, we aims to develop a more efficient and suitable TruncatedRPN
(TRPN) for lesion detection which can adaptively localize the lesion
object for different input.

Denote $\mathrm{\mathcal{I}}$ as the images feature extracted from
the input image, $(i,j)$ as specific location in $\mathrm{\mathrm{\mathcal{I}}}$,
original RPN can be formulate as $\mathcal{F}(\mathrm{\mathcal{I}}(i,j)|k)$,
$k$ is anchor prior related to location $(i,j)$ like scale and aspect
ratio settings. Following this formulation, TRPN introduces a module
to predict distribution of anchor localization $p(\mathrm{\mathcal{I}}(i,j))$
and truncates the localition whose predicted confidence are below
predefined threshold $\epsilon$, as follows:

\begin{equation}
TRPN=\mathcal{F}([p(\mathrm{\mathcal{I}}(i,j))\cdot\mathrm{\mathcal{I}}(i,j)]_{p(\mathrm{\mathcal{I}}(i,j))>\epsilon}|k)
\end{equation}

We empirically apply $1\times1$ convolution with an element-wise
sigmoid function as $p$. And we set mean of predicted distribution
as $\epsilon$. This mechanism can filter out half of the regions
and avoid redundant information.

\begin{table*}
\begin{centering}
\tabcolsep 0.015in{\footnotesize{}}%
\begin{tabular}{c|ccccccc|c|c|c}
\hline 
\multirow{2}{*}{{\scriptsize{}Method}} & \multicolumn{7}{c|}{{\scriptsize{}Sensitivity IoU (\%)}} & {\scriptsize{}mAP} & {\scriptsize{}Speed} & {\scriptsize{}\#.param}\tabularnewline
\cline{2-8} \cline{3-8} \cline{4-8} \cline{5-8} \cline{6-8} \cline{7-8} \cline{8-8} 
 & {\scriptsize{}0.5} & {\scriptsize{}1} & {\scriptsize{}2} & {\scriptsize{}3} & {\scriptsize{}4} & {\scriptsize{}6} & {\scriptsize{}8} & {\scriptsize{}(\%)} & {\scriptsize{}(img/ms)} & {\scriptsize{}(M)}\tabularnewline
\hline 
{\scriptsize{}3DCE, 3 slices \parencite{yan20183d}} & {\scriptsize{}56.49} & {\scriptsize{}67.65} & {\scriptsize{}76.89} & {\scriptsize{}80.56} & {\scriptsize{}82.76} & {\scriptsize{}85.71} & {\scriptsize{}87.03} & {\scriptsize{}25.6} & {\scriptsize{}42} & {\scriptsize{}17.2}\tabularnewline
{\scriptsize{}3DCE, 9 slices \parencite{yan20183d}} & {\scriptsize{}59.32} & {\scriptsize{}70.68} & {\scriptsize{}79.09} & {\scriptsize{}82.78} & {\scriptsize{}84.34} & {\scriptsize{}86.77} & {\scriptsize{}87.81} & {\scriptsize{}27.9} & {\scriptsize{}56} & {\scriptsize{}19.2}\tabularnewline
{\scriptsize{}3DCE, 27 slices \parencite{yan20183d}} & {\scriptsize{}62.48} & {\scriptsize{}73.37} & {\scriptsize{}80.70} & {\scriptsize{}83.61} & {\scriptsize{}85.65} & {\scriptsize{}87.26} & {\scriptsize{}89.09} & {\scriptsize{}29.2} & {\scriptsize{}114} & {\scriptsize{}25.2}\tabularnewline
\hline 
{\scriptsize{}FRCNN, 3 slices \parencite{ren2015faster}} & {\scriptsize{}55.94} & {\scriptsize{}66.25} & {\scriptsize{}74.69} & {\scriptsize{}79.17} & {\scriptsize{}79.87} & {\scriptsize{}79.87} & {\scriptsize{}79.87} & {\scriptsize{}27.9} & {\scriptsize{}54} & {\scriptsize{}33.0}\tabularnewline
{\scriptsize{}FPN, 3 slices \parencite{lin2017feature}} & {\scriptsize{}60.57} & {\scriptsize{}70.34} & {\scriptsize{}77.65} & {\scriptsize{}81.51} & {\scriptsize{}83.55} & {\scriptsize{}85.67} & {\scriptsize{}85.67} & {\scriptsize{}31.5} & {\scriptsize{}68} & {\scriptsize{}41.4}\tabularnewline
{\scriptsize{}Deformv2, 3 slices \parencite{zhu2018deformable}} & {\scriptsize{}63.16} & {\scriptsize{}72.35} & {\scriptsize{}81.04} & {\scriptsize{}84.47} & {\scriptsize{}86.52} & {\scriptsize{}87.76} & {\scriptsize{}87.76} & {\scriptsize{}34.3} & {\scriptsize{}74} & {\scriptsize{}42.2}\tabularnewline
\hline 
{\scriptsize{}ElixirNet, 3 slices} & {\scriptsize{}60.43$^{+4.49}$} & {\scriptsize{}69.38$^{+3.13}$} & {\scriptsize{}77.20$^{+2.51}$} & {\scriptsize{}80.52$^{+1.35}$} & {\scriptsize{}82.72$^{+2.85}$} & {\scriptsize{}83.74$^{+3.9}$} & {\scriptsize{}83.74$^{+3.9}$} & {\scriptsize{}30.8$^{+2.9}$} & {\scriptsize{}78} & {\scriptsize{}19.9}\tabularnewline
{\scriptsize{}ElixirNet w FPN, 3 slices} & \textbf{\scriptsize{}66.}{\scriptsize{}39$^{+5.82}$} & {\scriptsize{}75.41$^{+5.07}$} & {\scriptsize{}82.05$^{+4.4}$} & {\scriptsize{}86.63$^{+4.12}$} & {\scriptsize{}87.75$^{+4.2}$} & {\scriptsize{}88.36$^{+2.69}$} & {\scriptsize{}88.36$^{+2.69}$} & {\scriptsize{}35.2$^{+3.7}$} & {\scriptsize{}90} & {\scriptsize{}28.4}\tabularnewline
{\scriptsize{}ElixirNet w Deformv2, 3slices} & {\scriptsize{}65.64$^{+2.48}$} & \textbf{\scriptsize{}75.84$^{+3.49}$} & \textbf{\scriptsize{}83.70$^{+2.66}$} & \textbf{\scriptsize{}86.92}{\scriptsize{}$^{+2.45}$} & \textbf{\scriptsize{}88.91$^{+2.39}$} & \textbf{\scriptsize{}91.37$^{+3.61}$} & \textbf{\scriptsize{}91.37$^{+3.61}$} & \textbf{\scriptsize{}36.3$^{+2.0}$} & {\scriptsize{}102} & {\scriptsize{}29.3}\tabularnewline
\hline 
\end{tabular}{\footnotesize\par}
\par\end{centering}
\caption{\label{tab:sensitivity}Comparison with state-of-the-art object detection
on DeepLesion. Sensitivity at various FPs per image on the test set
of the official data split of DeepLesion. IoU criteria and 3 slices
input (one key slice and two neighbor slices ) are used.}
\end{table*}

\subsection{Auto-lesion Block}

The proposed Auto-lesion Block (ALB) is a medical data-friendly block
with hybrid receptive fields and kernel size. The Figure\ref{fig:cell}
shows the detailed architecture of ALB with keeping the same spatial
resolution and output channels with input, and ALB$_{r}$ with reducing
the half of spatial resolution and increasing double channels. In
ALB, dilated convolution with flexible receptive fields can locate
the tiny-size lesion, and Non-local operator captures relation-aware
information to distinguish lesions with similar no-lesion regions.
And ALB$_{r}$ is a inception-like block, mentioned in \cite{szegedy2016rethinking},
e.g. $3\times3$ convolution is composed in a asymmetric $1\times3$
and $3\times1$ convolution. It also uses large kernel depthwise convolution
and average pooling to keep a trade-off between computational overhead
and network complexity. These two blocks are customized by medical
data-oriented NAS with the same search space and strategy.

\subsubsection{Search Space for ALB}

Assume that ALB is a directed acyclic graph consisting of $\mathcal{B}$
branches, each branch has 2 inputs from previous branches and outputs
1 tensor following \cite{liu2018darts}, and initial channels of each
branch is \textbf{$D$}. Branch $b$ can be specified as a 5-tuple
($X_{1},X_{2},OP_{1},OP_{2},Y_{b}$), where $X_{1},X_{2}\in\mathcal{X}_{b}$
specific input tensors, $OP_{1},OP_{2}\in\mathcal{OP}$ specific operations
to apply to the corresponding inputs. The output $Y$ concatenates
with all branches outputs $Y=concat(Y_{1},Y_{2},...,Y_{b})$ with
$\mathcal{B}\times D$ channels. In most of the image classification
NAS frameworks \cite{zoph2016neural,zoph2018learning,liu2018darts},
similar candidate operations for $\mathcal{OP}$ are considered. These
pre-defined operations sets may be reasonable in the natural image
classification task, while medical lesion object detection depends
on semantic relation among proposals and flexible receptive fields.
In addition, \cite{liu2018receptive} show the relationship between
the size and eccentricity of receptive fields can enhance the discriminability
and robustness of the feature representation. Inspired by the above
methods, we consider 9 operators as the set of candidate operations
in $\mathcal{OP}$:
\begin{center}
{\footnotesize{}}%
\begin{tabular}{ll}
{\footnotesize{}$\circ$ no connection} & {\footnotesize{}$\circ$ $3\times3$ depthwise-separable conv}\tabularnewline
{\footnotesize{}$\circ$ skip connection} & {\footnotesize{}$\circ$ $5\times5$ depthwise-separable conv}\tabularnewline
{\footnotesize{}$\circ$ $3\times3$ average pooling} & {\footnotesize{}$\circ$ $3\times3$ atrous conv w dilate rate
3}\tabularnewline
{\footnotesize{}$\circ$ Non-local} & {\footnotesize{}$\circ$ $3\times3$ atrous conv w dilate rate
5}\tabularnewline
\multicolumn{2}{l}{{\footnotesize{}$\circ$ $1\times3$ and $3\times1$ depthwise-separable
conv}}\tabularnewline
\end{tabular}{\footnotesize\par}
\par\end{center}

All operations are of stride 1 and the convolved feature maps are
padded to preserve their spatial resolution. We use the ReLU-Conv-BN
order for convolutional operations. Our search space contains $3\times10^{-9}$
structures.

\textbf{Non-local Operator.} The Non-local operator aims to encode
semantic relation between region proposals which is relevant to the
object detection. We formulate the relation as a region-to-region
undirected graph $\boldsymbol{G}$ : $\boldsymbol{\boldsymbol{G}}=<\mathcal{N},\mathcal{\mathcal{E}}>$,
where each node in $\mathcal{N}$ corresponds to a region proposal
and each edge $e_{i,j}\in\mathcal{E}$ encodes relationship between
two nodes. Formally, the input of non-local operator is the proposal
features $X\in\mathbb{R}^{\mathcal{N}\times D}$ from previous branches.
The adjacency matrix for the undirected graph $\boldsymbol{G}$ with
self-loops can be then calculated by a matrix multiplication as:

\begin{equation}
\mathcal{E}=softmax(\phi(X)\phi(X)^{T}),\label{eq:edge}
\end{equation}

where $\phi(.)$ is non-linear transformation with ReLU activation.
Then we use a layer GCN \parencite{kipf2016semi} for the propagation
of each node in the graph $\mathcal{E}$. A simple form is $Y=\sigma(\mathcal{E}\psi(X)W)$,
where $W$ and $\psi(.)$ are non-linear transformation, and $\sigma$
is active function. In this paper, we consider a fully-connected layer
with ReLU activation as the non-linear function and output dimension
is $\frac{D}{2}$. For the output of this operation, we use another
fully-connected layer $g$ to keep the shape of the input tensor.
The final output of the Non-local operator is $g(Y)$.

\subsubsection{Differentiable Search Strategy}

The search strategy of ALB builds on a continuous relaxation of the
discrete architectures described in \cite{liu2018darts}. The branch\textquoteright s
output tensor $Y_{i}$ is a weighted mixture of the operations with
$|\mathcal{OP}|$ parallel paths which are connected to all elements
in $X_{i}$:

\begin{equation}
Y_{i}=\sum_{Y_{j}\in X_{i}}\sum_{\mathcal{OP}_{k}\in\mathcal{OP}}\alpha_{ij}^{k}\mathcal{OP}_{k}(Y_{j})
\end{equation}

where the weight $\left\{ \alpha_{ij}^{k}\right\} $ are the architecture
parameters calculated by applying softmax to $|\mathcal{OP}|$, $k$is
in $\{ALB,ALB_{r}\}$. After the continuous relaxation, the task of
architecture search reduces to learning a set of continuous variables
$\left\{ \alpha_{ij}^{k}\right\} $. Therefore they can be optimized
efficiently using stochastic gradient descent (SGD). We randomly split
training data into two disjoint sets, and the loss function includes
cross-entropy about classification and smooth $L1$ loss about location,
calculated on these sets are denoted by $\mathcal{L}_{train}$ and
$\mathcal{L}_{val}$. The optimization alternates between:
\begin{enumerate}
\item Update network weights $\omega$ by $\nabla_{\omega}\mathcal{L}_{train}(\omega,\alpha)$
\item Update architecture weights $\alpha$ by $\nabla_{\alpha}\mathcal{L}_{val}(\omega,\alpha)$
\end{enumerate}
At the end of search, a discrete architecture is obtained by replacing
each mixed operation $\mathcal{OP}_{ij}$ with the most likely operation
by argmax: $\mathcal{OP}_{ij}=argmax_{\mathcal{OP}_{k}\in\mathcal{OP}}\alpha_{ij}^{k}$.

\begin{table*}
\begin{centering}
\tabcolsep 0.015in
\par\end{centering}
\begin{centering}
\tabcolsep 0.015in{\footnotesize{}}%
\begin{tabular}{c|ccccccc}
\hline 
\multirow{2}{*}{{\scriptsize{}Method}} & \multicolumn{7}{c}{{\scriptsize{}Sensitivity IoBB (\%)}}\tabularnewline
\cline{2-8} \cline{3-8} \cline{4-8} \cline{5-8} \cline{6-8} \cline{7-8} \cline{8-8} 
 & {\scriptsize{}0.5} & {\scriptsize{}1} & {\scriptsize{}2} & {\scriptsize{}3} & {\scriptsize{}4} & {\scriptsize{}6} & {\scriptsize{}8}\tabularnewline
\hline 
{\scriptsize{}3DCE, 3 slices \parencite{yan20183d}} & {\scriptsize{}58.43} & {\scriptsize{}70.95} & {\scriptsize{}80.64} & {\scriptsize{}85.12} & {\scriptsize{}87.30} & {\scriptsize{}90.72} & {\scriptsize{}92.37}\tabularnewline
{\scriptsize{}3DCE, 9 slices \parencite{yan20183d}} & {\scriptsize{}62.64} & {\scriptsize{}74.37} & {\scriptsize{}83.29} & {\scriptsize{}86.99} & {\scriptsize{}89.15} & {\scriptsize{}91.61} & {\scriptsize{}92.90}\tabularnewline
{\scriptsize{}3DCE, 27 slices \parencite{yan20183d}} & {\scriptsize{}64.01} & {\scriptsize{}75.69} & {\scriptsize{}83.71} & {\scriptsize{}87.52} & {\scriptsize{}88.25} & {\scriptsize{}91.47} & {\scriptsize{}92.85}\tabularnewline
\hline 
{\scriptsize{}FRCNN, 3 slices \parencite{ren2015faster}} & {\scriptsize{}57.59} & {\scriptsize{}68.46} & {\scriptsize{}77.36} & {\scriptsize{}81.96} & {\scriptsize{}82.43} & {\scriptsize{}82.43} & {\scriptsize{}82.43}\tabularnewline
{\scriptsize{}FPN, 3 slices \parencite{lin2017feature}} & {\scriptsize{}62.17} & {\scriptsize{}72.35} & {\scriptsize{}80.29} & {\scriptsize{}83.82} & {\scriptsize{}86.03} & {\scriptsize{}87.68} & {\scriptsize{}87.68}\tabularnewline
{\scriptsize{}Deform, 3 slices \parencite{zhu2018deformable}} & {\scriptsize{}65.09} & {\scriptsize{}74.49} & {\scriptsize{}83.34} & {\scriptsize{}86.84} & {\scriptsize{}88.86} & {\scriptsize{}89.71} & {\scriptsize{}89.71}\tabularnewline
\hline 
{\scriptsize{}ElixirNet, 3 slices} & {\scriptsize{}63.36$^{+5.77}$} & {\scriptsize{}73.21$^{+4.75}$} & {\scriptsize{}81.38$^{+4.02}$} & {\scriptsize{}85.06$^{+3.10}$} & {\scriptsize{}86.53$^{+4.10}$} & {\scriptsize{}87.50$^{+5.07}$} & {\scriptsize{}87.50$^{+5.07}$}\tabularnewline
{\scriptsize{}ElixirNet w FPN, 3 slices} & \textbf{\scriptsize{}68.67}{\scriptsize{}$^{+6.50}$} & \textbf{\scriptsize{}77.93}{\scriptsize{}$^{+5.58}$} & {\scriptsize{}84.83$^{+4.54}$} & {\scriptsize{}88.62$^{+4.80}$} & {\scriptsize{}90.59$^{+4.56}$} & {\scriptsize{}91.77$^{+4.09}$} & {\scriptsize{}91.77$^{+4.09}$}\tabularnewline
{\scriptsize{}ElixirNet w Deformv2, 3slices} & {\scriptsize{}67.33$^{+2.24}$} & {\scriptsize{}77.89$^{+3.40}$} & \textbf{\scriptsize{}86.00$^{+2.66}$} & \textbf{\scriptsize{}90.13$^{+3.29}$} & \textbf{\scriptsize{}91.66$^{+2.80}$} & \textbf{\scriptsize{}93.55$^{+3.84}$} & \textbf{\scriptsize{}93.55$^{+3.84}$}\tabularnewline
\hline 
\end{tabular}{\footnotesize\par}
\par\end{centering}
\caption{\label{tab:sensitivity-iobb}Sensitivity with IoBB overlap criteria
at various FPs per image on the test set of the official data split
of DeepLesion.}
\end{table*}

\subsection{Relation Transfer Module}

To develop a universal multi-purpose CAD framework, we introduce Relation
Transfer Module (RTM) to endow multi-type lesion detection. Rather
than simply finetuning a new multi-type classifier, we transfer coarse
embeddings of lesion existing to fine lesion types information and
try to propagate semantic embeddings among proposals by learned relation
$\mathcal{\mathcal{E}}$ in ALB.

\begin{table}[t]
\begin{centering}
\tabcolsep 0.015in{\footnotesize{}}%
\begin{tabular}{c|ccc|c}
\hline 
\multirow{2}{*}{{\scriptsize{}Method}} & \multicolumn{3}{c|}{{\scriptsize{}Sensitivity IoU (\%)}} & \multirow{2}{*}{{\scriptsize{}mAP (\%)}}\tabularnewline
\cline{2-4} \cline{3-4} \cline{4-4} 
 & {\scriptsize{}0.5} & {\scriptsize{}1} & {\scriptsize{}2} & \tabularnewline
\hline 
{\scriptsize{}FPN \parencite{lin2017feature}} & {\scriptsize{}53.68} & {\scriptsize{}56.71} & {\scriptsize{}58.46} & {\scriptsize{}21.1}\tabularnewline
{\scriptsize{}Deformv2 \parencite{zhu2018deformable}} & {\scriptsize{}67.18} & {\scriptsize{}68.31} & {\scriptsize{}71.21} & {\scriptsize{}32.9}\tabularnewline
\hline 
{\scriptsize{}ElixirNet w FPN} & {\scriptsize{}54.69}\textbf{\scriptsize{}$^{+1.01}$} & {\scriptsize{}57.93$^{+1.22}$} & {\scriptsize{}61.20$^{+2.74}$} & {\scriptsize{}29.6$^{+8.5}$}\tabularnewline
{\scriptsize{}ElixirNet w Deformv2} & \textbf{\scriptsize{}67.28$^{+0.10}$} & \textbf{\scriptsize{}69.02$^{+0.71}$} & \textbf{\scriptsize{}80.05$^{+2.4}$} & \textbf{\scriptsize{}33.6$^{+0.7}$}\tabularnewline
\hline 
\end{tabular}{\footnotesize\par}
\par\end{centering}
\caption{\label{tab:gen}Comparison on Kits19 with 3 slices input.}
\end{table}

In some zero/few-shot problems, \cite{salakhutdinov2011learning,gong2018frage}
use the classifier\textquoteright s weights as the high-level semantic
embedding or representation of category. Thus the collection of the
weights $M\in\mathbb{R}^{2\times(P+1)}$ of original binary classifier
(including the bias) is regarded as category-wise semantic embeddings,
$P$ is the output dimension of the ALB$_{r}$. Since our graph $G$
is a region-to-region graph extracted from Non-local operator in ALB,
we need to find most appropriate mappings from category-wise semantic
embeddings to region-wise representations of node $f_{i}\in\mathsf{f}$
(the input of RTM). For avoiding bias produced by original binary
classification, we use a soft-mapping which compute the mapping weights
$\gamma_{M\rightarrow x_{i}}\in\mathbf{\Gamma}^{s}$ as $\gamma_{M\rightarrow x_{i}}=\frac{\exp(s_{ij})}{\sum_{j}\exp(s_{ij})}$,
where $s_{ij}$ is the classification score for the region $i$ towards
category $j$ from the previous binary classification layer, denoted
``FC 2-cls'' in Figure \ref{fig:framework}. Then the process of
graph reasoning can be solved by matrix multiplication:
$\mathsf{f}_{E}=\mathcal{E}\Gamma^{s}MW_{E},$
where $W_{E}\in\mathbb{R^{\mathcal{N}\times E}}$ is a transformation
weight matrix and $E$ is the output dimension of the RTM module.
Finally, the enhanced feature $\mathsf{f}_{E}$ is concatenated to
the original region features $\mathsf{f}$ to improve both classification
and localization of the multi-type lesion.

\begin{table}[t]
\begin{centering}
\tabcolsep 0.015in{\footnotesize{}}%
\begin{tabular}{c|c|c|c|c|c|c|c|c}
\hline 
\multirow{2}{*}{{\tiny{}\%}} & \multicolumn{8}{c}{{\tiny{}Lesion Type}}\tabularnewline
\cline{2-9} \cline{3-9} \cline{4-9} \cline{5-9} \cline{6-9} \cline{7-9} \cline{8-9} \cline{9-9} 
 & \multicolumn{1}{c|}{{\tiny{}BN}} & \multicolumn{1}{c|}{{\tiny{}AB}} & \multicolumn{1}{c|}{{\tiny{}ME}} & \multicolumn{1}{c|}{{\tiny{}LV}} & \multicolumn{1}{c|}{{\tiny{}LU}} & \multicolumn{1}{c|}{{\tiny{}KD}} & \multicolumn{1}{c|}{{\tiny{}ST}} & \multicolumn{1}{c}{{\tiny{}PV}}\tabularnewline
\hline 
{\tiny{}FPN w finetune \parencite{lin2017feature}} & {\tiny{}-} & {\tiny{}79.58} & {\tiny{}71.34} & {\tiny{}78.05} & {\tiny{}83.32} & {\tiny{}-} & {\tiny{}57.96} & {\tiny{}72.21}\tabularnewline
{\tiny{}ElixirNet w/o RTM} & {\tiny{}56.52} & {\tiny{}84.14} & {\tiny{}83.41} & {\tiny{}82.99} & {\tiny{}87.10} & {\tiny{}60.70} & {\tiny{}75.67} & {\tiny{}80.64}\tabularnewline
{\tiny{}ElixirNet} & \textbf{\tiny{}57.38} & \textbf{\tiny{}85.41} & \textbf{\tiny{}84.62} & \textbf{\tiny{}83.94} & \textbf{\tiny{}88.72} & \textbf{\tiny{}65.51} & \textbf{\tiny{}77.24} & \textbf{\tiny{}83.07}\tabularnewline
\hline 
\end{tabular}{\footnotesize\par}
\par\end{centering}
\caption{\label{tab:weakly-1}Multi-type lesion detection of sensitivity at
3 FPs per image on the test set. The abbreviations of lesion types
stand for bone, abdomen, mediastinum, liver, lung, kidney, soft tissue,
pelvis and bone, respectively. IoU as overlap computation criteria
is used. \textquotedblleft FPN w finetune\textquotedblright{} and
``ElixirNet w/o RTM'' finetune new multi-type classifier on validation
datasets. ``-'' means too low recall to evaluate sensitivity.}
\end{table}

\section{Experiments}

\textbf{Datasets and Evaluations.} We conduct experiments on the DeepLesion \parencite{yan2018deeplesion}
and Kits19 \parencite{heller2019kits19} datasets. DeepLesion is a
large-scale dataset on 876,934 axial CT slices (mostly $512\times512$)
from 10,594 CT studies of 4,427 unique patients. DeepLesion contains
a variety of lesions \cite{yan2018deep}. However, only 32,120 axial
slices have bounding boxes annotation. There are 1\textendash 3 lesions
in each axial slice, totally 32,735 lesions. We select 30\% as validation
(4889 lesions) and test (4927 lesions), while the rest is regarded
as the training set (22919 images) following the official division
\cite{yan20183d,yan2018deeplesion}. The lesions in validation and
test sets are all labeled with specific types, and have been categorized
into the 8 subtypes of lung (2394 lesions), abdomen (2176 lesions),
mediastinum (1672 lesions), liver (1284 lesions), pelvis (681 lesions),
soft tissue (867 lesions), kidney (495 lesions), and bone (247 lesions).
The lesions in the training set are annotated by bounding box but
without labels of lesion types. To validate the generalization capability
of the model, we conduct experiments on Kits19 \parencite{heller2019kits19}
which is a kidney tumor semantic segmentation challenge. We random
split 80\% patients (totally 210 kidney cancer patients) as our training
set (13147 images) and the left as testing set (3209 images). Since
Kits19 is a segmentation dataset, we convert segmentation masks to
bounding boxes for all lesion.

\begin{figure*}[t]
\begin{centering}
\includegraphics[scale=0.34]{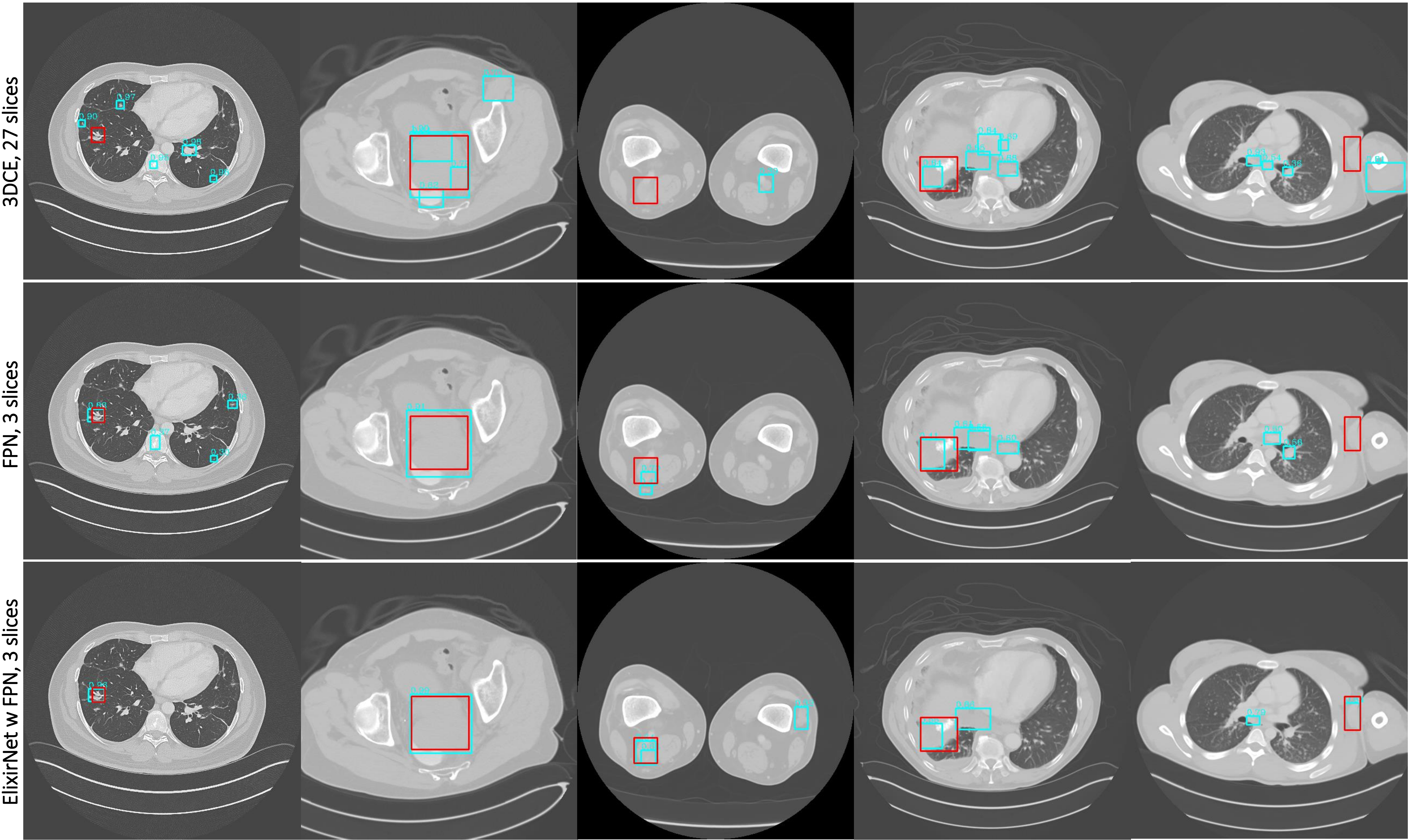}
\par\end{centering}
\caption{\label{fig:quality-results}Qualitative results on DeepLesion among
3DCE with 27 slices, FPN and our ElixirNet with FPN. Detection results
with higher confidence, precise location and less false positive can
be achieved by our method. The prediction with confidence score \textgreater{}
0.3 is visualized. The location of ground-truth (red box) is shown
for easy identification.}
\end{figure*}

For all the evaluation, we adopt mean Average Precision (mAP) across
IoU 0.5-0.9 as evaluation, which is designed for natural images detection
following \cite{lin2017feature}. Sensitivity \textbf{$(\mathrm{IoU}\geq0.5)$}
with different false positives per image is a commonly used metric
in medical detection \cite{yan20183d}. Considering the prediction
may still be viewed as a true positive (TP), it can also help the
radiologists when IoU is less than 0.5. To overcome this evaluation
bias, we also utilized the intersection over the detected bounding-box
area ratio (IoBB) as another criterion, following \cite{yan20183d}.

\textbf{Implementation Details.} We conduct all experiments on a single server with 4 GTX1080 cards
in Pytorch \parencite{paszke2017automatic}. Three widely-adopted
natural images detection methods i.e. FRCNN \parencite{ren2015faster},
FPN \parencite{lin2017feature} and Deformv2{\footnotesize{} \parencite{zhu2018deformable}}
are regarded as baseline network to show the generalization ability
of our ElixirNet. Unless otherwise noted, settings are the same for
all experiments.

\textbf{TruncatedRPN. }Based on the original RPN, we empirically apply
$1\times1$ convolution with an element-wise sigmoid function on RPN
feature map to obtaining the confidence map of the locations. Region
proposals are then generated at the locations where the confidence
is greater than the mean of the confidence map. Five anchor scales
$(2,3,4,6,12)$ and three aspect ratios of $(0.5,1,2)$ are used in
TRPN following \cite{yan20183d}.

\textbf{Auto-Lesion Block Search. }Owing to memory restriction, feature
extractor and RPN are frozen with initializing pretrained baseline
network on DeepLesion. We consider the branch $\mathcal{B}=4$ and
the initial channels $D$ of each branch is 16. To carry out architecture
search, we split half of the DeepLesion training data as the \textit{val}
set. Network weights $\omega$is updated after training 15 epochs
with batch size $=$16 (same with validation sets). We choose momentum
SGD with initial learning rate 0.02 (annealed down to zero following
a cosine schedule), momentum 0.9, and weight decay $3\times10^{-4}$
as an optimizer for weights $\omega$, while architecture weights
\textgreek{a} are optimized by Adam with initial learning rate $3\times10^{-4}$
, momentum 0.999 and weight decay $10^{-3}$.

\textbf{ElixirNet. }Series connection of feature extractor, TRPN,
1 ALB, 2 ALB$_{r}$ and RTM forms our final ElixirNet. The detailed
architecture of ALB and ALB$_{r}$ can be found in Figure \ref{fig:cell}
and we double initial channels $D$ in Auto-Lesion Block Search as
channels of each operator, and use a $3\times3$ convolution to reduce
output feature channels to 128 before feeding to ALB. And we set $P=512,E=128$
in RTM. For network training, stochastic gradient descent (SGD) is
performed on 4 GPUs per 2 images for 12 epochs. ResNet-50 \parencite{he2016deep}
pretrained on ImageNet \cite{russakovsky2015imagenet} is used. The
initial learning rate is 0.02, reduce three times ($\times0.01$);
$10^{-4}$ as weight decay; 0.9 as momentum.

\begin{figure}
\begin{centering}
\includegraphics[scale=0.34]{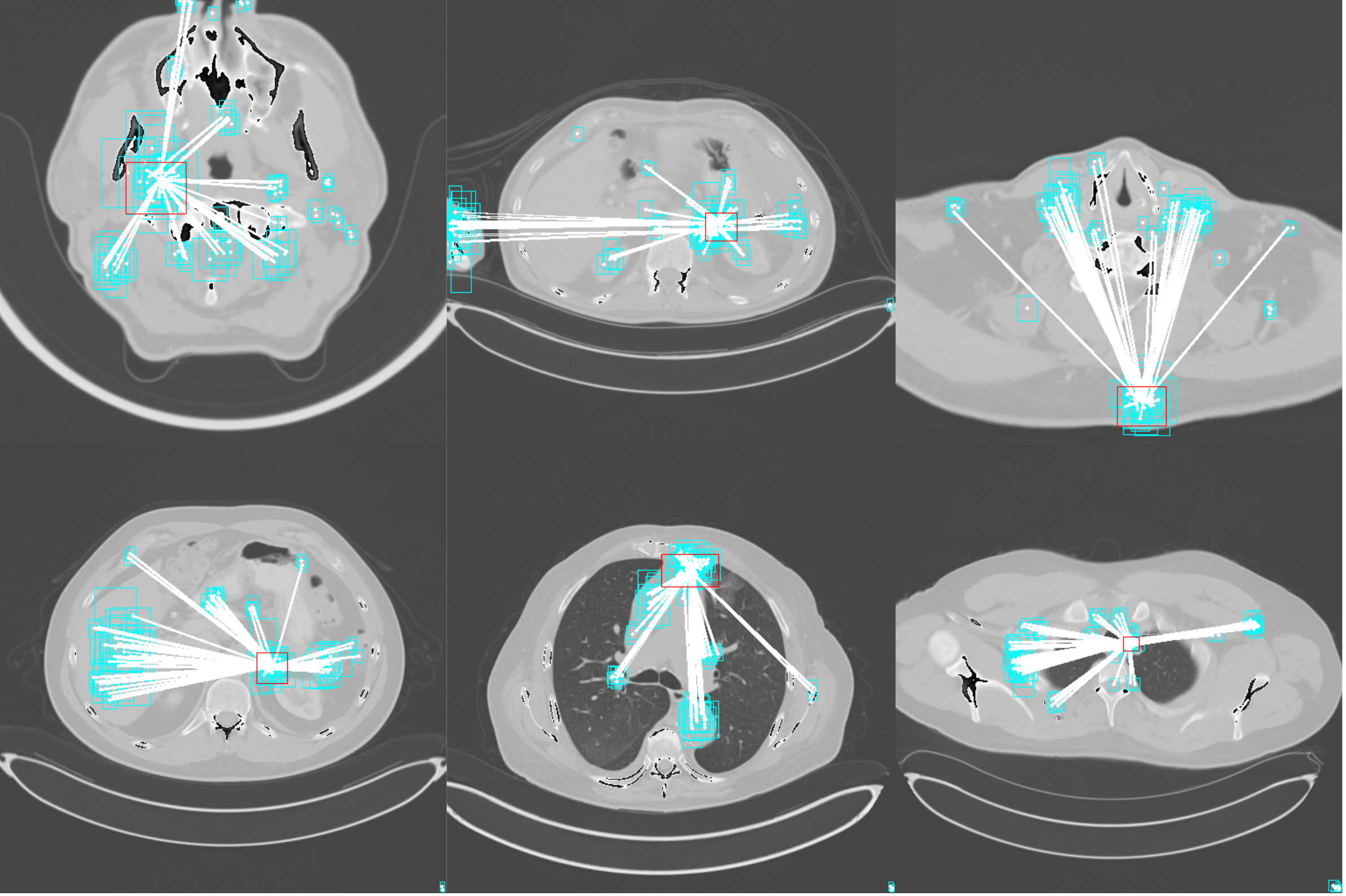}
\par\end{centering}
\caption{\label{fig:nonlocal_edges} Visualization of learned relation $G$
from ALB. The centers of regions are plotted and connected by the
learned edges $\mathcal{E}$. Edges thickness correspond to the strength
of the graph edge weights. Note other regions are linked to regions
which are close to ground-truth (red box) location.}
\end{figure}

\subsection{Comparison with state-of-the-art}

\textbf{Single Lesion Detection Benchmarks.} The results of Lesion
Detection are presented in Table \ref{tab:sensitivity} and Table
\ref{tab:sensitivity-iobb}. We compare with 3DCE \parencite{yan20183d},
FRCNN \parencite{ren2015faster}, FPN \parencite{lin2017feature}
and Deformv2 \parencite{zhu2018deformable} with FPN backbone. Notably,
ElixirNet achieved significant improvements than all natural images
baselines with fewer parameters. Our ElixirNet improves around 10.3\%
for FRCNN, 11.7\% for FPN and 5.8\% for Deformv2 on mAP respectively.
And the result shows ElixirNet works mostly on small false positive
per images (improving 5.82\% of IoU overlap 6.50\% of IoBB overlap
sensitivity at 0.5 FPs on FPN). Our method with high accuracy at low
error rates is consistent with the needs of radiologists. The parameter
size of ElixirNet is about 60\%-70\% of the baseline detection network.
Without optimization of parallel structure in Pytorch, inference speed
is slightly slower than baseline. We also evaluate Kits19\parencite{heller2019kits19}
to compare with FPN \parencite{lin2017feature} and Deformv2 \parencite{zhu2018deformable}
with FPN backbone. The results can be found in Table \ref{tab:gen}.
As can be seen, our method outperforms 40.3\% and 2\% than FPN and
Deformv2 on mAP, and has to continue improvements on sensitivity with
IoU. Figure \ref{fig:quality-results} shows quality results comparison
of the 3DCE with 27 slices input, FPN and our ElixirNet with FPN on
DeepLesion. Our method is more accurate than the 3DCE and baseline
FPN due to relation-aware architecture adaptation for medical images.
The graph structure learned from ALB is in Figure \ref{fig:nonlocal_edges}.
The proposed ALB learned interpretable relation among regions. Other
regions are linked to regions which are close to the ground-truth
location. The relation-aware knowledge helps improve proposals feature
thus leading to better detection performance.\textbf{ Multi-type Lesion
Detection. }Different lesion types of labelling are provided only
in the test and validation set of DeepLesion \cite{yan2018deeplesion}.
A simple solution is replacing binary classifier and bounding-box
regression with new multi-type classifier and bounding-box regression,
denoted as ``FPN w finetune'' and ``ElixirNet w/o RTM''. During
the training phase, feature extractor and RPN are frozen, while RCNN
head and our ALB finetune for 6 epochs with 0.002 learning rate on
the validation set. The results in Table \ref{tab:weakly-1} show
that our ``ElixirNet w/o RTM'' far outperforms FPN on generalization
and robustness of multi-type lesion detection. Furthermore, the results
with RTM achieve an average 2\% sensitivity of all types with only
increasing 0.2\% parameters.

\begin{table}
\begin{centering}
\tabcolsep 0.015in{\footnotesize{}}%
\begin{tabular}{c|ccc|c|c|c}
\hline 
\multirow{2}{*}{{\scriptsize{}\%}} & \multirow{2}{*}{{\scriptsize{}TRPN}} & \multirow{2}{*}{{\scriptsize{}ALB}} & \multirow{2}{*}{{\scriptsize{}RTM}} & {\scriptsize{}input} & {\scriptsize{}3 FPs (IoU)} & \multirow{2}{*}{{\scriptsize{}mAP}}\tabularnewline
 &  &  &  & {\scriptsize{}9 slices} & {\scriptsize{}sensitivity} & \tabularnewline
\hline 
{\scriptsize{}FPN \parencite{lin2017feature}} &  &  &  &  & {\scriptsize{}81.51} & {\scriptsize{}31.5}\tabularnewline
\hline 
 & \textbf{\scriptsize{}$\surd$} &  &  &  & {\scriptsize{}83.61} & {\scriptsize{}33.9}\tabularnewline
 &  & \textbf{\scriptsize{}$\surd$} &  &  & {\scriptsize{}83.88} & {\scriptsize{}34.1}\tabularnewline
 &  & \textbf{\scriptsize{}$\surd$} &  & \textbf{\scriptsize{}$\surd$} & {\scriptsize{}84.19} & {\scriptsize{}34.0}\tabularnewline
 & \textbf{\scriptsize{}$\surd$} & \textbf{\scriptsize{}$\surd$} &  &  & {\scriptsize{}84.72} & {\scriptsize{}34.8}\tabularnewline
 & \textbf{\scriptsize{}$\surd$} &  & \textbf{\scriptsize{}$\surd$} &  & {\scriptsize{}85.43} & {\scriptsize{}34.4}\tabularnewline
\hline 
{\scriptsize{}ElixirNet} & \textbf{\scriptsize{}$\surd$} & \textbf{\scriptsize{}$\surd$} & \textbf{\scriptsize{}$\surd$} &  & {\scriptsize{}86.63} & {\scriptsize{}35.2}\tabularnewline
\hline 
\end{tabular}{\footnotesize\par}
\par\end{centering}
\caption{\label{tab:Ablation-Studies-on}Ablation Studies on sensitivity and
mAP. The impact of different modules and input slices is explored.}
\end{table}

\begin{table}[t]
\begin{centering}
\tabcolsep 0.015in{\footnotesize{}}%
\begin{tabular}{c|c|c}
\hline 
\multirow{1}{*}{{\footnotesize{}Method and Modifications}} & {\footnotesize{}sensitivity (\%)} & {\footnotesize{}mAP (\%)}\tabularnewline
\hline 
{\footnotesize{}ElixirNet} & {\footnotesize{}83.88} & {\footnotesize{}34.1}\tabularnewline
\hline 
{\footnotesize{}ALB w/o our search space} & \multirow{2}{*}{{\footnotesize{}81.06$^{-2.82}$}} & \multirow{2}{*}{{\footnotesize{}32.1$^{-2.0}$}}\tabularnewline
{\footnotesize{}\parencite{liu2018darts}} &  & \tabularnewline
\hline 
{\footnotesize{}random ALB \parencite{li2019random}} & {\footnotesize{}81.84$^{-2.04}$} & {\footnotesize{}32.1$^{-2.0}$}\tabularnewline
\hline 
\end{tabular}{\footnotesize\par}
\par\end{centering}
\caption{\label{tab:alb}Comparison of different search space and strategy.
The input with 3 slices and evaluation on sensitivity at 3 FPs (IoU
criteria) and mAP are used for all methods. ``w/o our search space''
means using the search space in image classification NAS works directly.
``random ALB'' replaces random search strategy with our differentiable
search strategy.}
\end{table}

\textbf{Ablative Analysis.} To analyze the importance of different modules in Table \ref{tab:Ablation-Studies-on},
ablation studies are conducted on FPN {\footnotesize{}\parencite{lin2017feature}}
baseline. All results are on sensitivity with IoU overlap criteria
at 3 FPs per image and mAP. 1) ALB is the most vital component with
relation-aware adaptation for medical images. 2) TRPN can improve
performance for false positive reduction and outperforms focal loss
\cite{lin2017focal} with replacing the cross-entropy loss in RPN
whose sensitivity at 3 FPs (with IoU criteria) is 82.66\%. 3) The
usage of RTM further improves the sensitivity by 1.91\% and mAP by
0.4\%. 4) More slices are considered following \cite{yan20183d};
the performance of the 9 slices is slightly higher than the 3 slices.
Since 72\% of lesions can be covered by 3 slices input in DeepLesion
dataset, our final model uses 3 slices input. \textbf{Search space
and strategy in ALB.} The comparison from Table \ref{tab:alb} shows
that the proposed novel search space is most significant for designing
specific architecture for medical images. Original search space in
previous NAS work causes 2.82\% sensitivity falloff and 2\% decrease
on mAP. Also, differentiable search strategy searching architectures
based on the gradient descent is a benefit for medical lesion detection.

\section{Conclusion}

In this work, we proposed a novel ElixirNet which is customized for
medical lesion detection with a composition of TruncatedRPN, Auto-lesion
Block and Relation transfer module. It can adaptively suppress anchor
location with interested lesion absence and captures the semantic
context of a key proposal from relation-aware neighbourhoods, leading
to more suitable and efficient prediction and false positive reduction.
The stable and consistent performance of our ElixirNet on all evaluation
criteria of DeepLesion and Kits19 outperforms current methods with
fewer parameters.\textbf{ }

\textbf{Acknowledgments.} This research is partly supported by the
National Natural Science Foundation of China (No. U1611461) and in
part by the National Natural Science Foundation of China (NSFC) under
Grant No.61976233.

\fontsize{9.0pt}{10.0pt} \selectfont
\bibliographystyle{aaai} \bibliography{4006}

\begin{thebibliography}{}

\bibitem[\protect\citeauthoryear{Brock \bgroup et al\mbox.\egroup
  }{2017}]{brock2017smash}
Brock, A.; Lim, T.; Ritchie, J.~M.; and Weston, N.
\newblock 2017.
\newblock Smash: one-shot model architecture search through hypernetworks.
\newblock {\em arXiv preprint arXiv:1708.05344}.

\bibitem[\protect\citeauthoryear{Chen \bgroup et al\mbox.\egroup
  }{2019}]{chen2019detnas}
Chen, Y.; Yang, T.; Zhang, X.; Meng, G.; Pan, C.; and Sun, J.
\newblock 2019.
\newblock Detnas: Neural architecture search on object detection.
\newblock In {\em NeurIPS}.

\bibitem[\protect\citeauthoryear{Dai \bgroup et al\mbox.\egroup
  }{2016}]{dai2016r}
Dai, J.; Li, Y.; He, K.; and Sun, J.
\newblock 2016.
\newblock R-fcn: Object detection via region-based fully convolutional
  networks.
\newblock In {\em NerIPS}.

\bibitem[\protect\citeauthoryear{de Moor \bgroup et al\mbox.\egroup
  }{2018}]{de2018automated}
de~Moor, T.; Rodriguez-Ruiz, A.; M{\'e}rida, A.~G.; Mann, R.; and Teuwen, J.
\newblock 2018.
\newblock Automated soft tissue lesion detection and segmentation in digital
  mammography using a u-net deep learning network.
\newblock In {\em IWBI}.

\bibitem[\protect\citeauthoryear{Ghiasi, Lin, and Le}{2019}]{ghiasi2019fpn}
Ghiasi, G.; Lin, T.-Y.; and Le, Q.~V.
\newblock 2019.
\newblock Nas-fpn: Learning scalable feature pyramid architecture for object
  detection.
\newblock In {\em CVPR}.

\bibitem[\protect\citeauthoryear{Gong \bgroup et al\mbox.\egroup
  }{2018}]{gong2018frage}
Gong, C.; He, D.; Tan, X.; Qin, T.; Wang, L.; and Liu, T.-Y.
\newblock 2018.
\newblock Frage: Frequency-agnostic word representation.
\newblock In {\em NerIPS}.

\bibitem[\protect\citeauthoryear{Greenspan, Van~Ginneken, and
  Summers}{2016}]{greenspan2016guest}
Greenspan, H.; Van~Ginneken, B.; and Summers, R.~M.
\newblock 2016.
\newblock Guest editorial deep learning in medical imaging: Overview and future
  promise of an exciting new technique.
\newblock {\em IEEE Transactions on Medical Imaging} 35(5):1153--1159.

\bibitem[\protect\citeauthoryear{He \bgroup et al\mbox.\egroup
  }{2016}]{he2016deep}
He, K.; Zhang, X.; Ren, S.; and Sun, J.
\newblock 2016.
\newblock Deep residual learning for image recognition.
\newblock In {\em CVPR}.

\bibitem[\protect\citeauthoryear{Heller \bgroup et al\mbox.\egroup
  }{2019}]{heller2019kits19}
Heller, N.; Sathianathen, N.; Kalapara, A.; Walczak, E.; Moore, K.; Kaluzniak,
  H.; Rosenberg, J.; Blake, P.; Rengel, Z.; Oestreich, M.; et~al.
\newblock 2019.
\newblock The kits19 challenge data: 300 kidney tumor cases with clinical
  context, ct semantic segmentations, and surgical outcomes.
\newblock {\em arXiv preprint arXiv:1904.00445}.

\bibitem[\protect\citeauthoryear{Hu \bgroup et al\mbox.\egroup
  }{2018}]{hu2018relation}
Hu, H.; Gu, J.; Zhang, Z.; Dai, J.; and Wei, Y.
\newblock 2018.
\newblock Relation networks for object detection.
\newblock In {\em CVPR}.

\bibitem[\protect\citeauthoryear{Jaeger \bgroup et al\mbox.\egroup
  }{2018}]{jaeger2018retina}
Jaeger, P.~F.; Kohl, S.~A.; Bickelhaupt, S.; Isensee, F.; Kuder, T.~A.;
  Schlemmer, H.-P.; and Maier-Hein, K.~H.
\newblock 2018.
\newblock Retina u-net: Embarrassingly simple exploitation of segmentation
  supervision for medical object detection.
\newblock {\em arXiv preprint arXiv:1811.08661}.

\bibitem[\protect\citeauthoryear{Kipf and Welling}{2017}]{kipf2016semi}
Kipf, T.~N., and Welling, M.
\newblock 2017.
\newblock Semi-supervised classification with graph convolutional networks.
\newblock In {\em ICLR}.

\bibitem[\protect\citeauthoryear{Krizhevsky, Sutskever, and
  Hinton}{2012}]{krizhevsky2012imagenet}
Krizhevsky, A.; Sutskever, I.; and Hinton, G.~E.
\newblock 2012.
\newblock Imagenet classification with deep convolutional neural networks.
\newblock In {\em NerIPS}.

\bibitem[\protect\citeauthoryear{Li and Talwalkar}{2019}]{li2019random}
Li, L., and Talwalkar, A.
\newblock 2019.
\newblock Random search and reproducibility for neural architecture search.
\newblock {\em arXiv preprint arXiv:1902.07638}.

\bibitem[\protect\citeauthoryear{Liao \bgroup et al\mbox.\egroup
  }{2019}]{liao2019evaluate}
Liao, F.; Liang, M.; Li, Z.; Hu, X.; and Song, S.
\newblock 2019.
\newblock Evaluate the malignancy of pulmonary nodules using the 3-d deep leaky
  noisy-or network.
\newblock {\em IEEE transactions on neural networks and learning systems}.

\bibitem[\protect\citeauthoryear{Lin \bgroup et al\mbox.\egroup
  }{2017a}]{lin2017feature}
Lin, T.-Y.; Doll{\'a}r, P.; Girshick, R.; He, K.; Hariharan, B.; and Belongie,
  S.
\newblock 2017a.
\newblock Feature pyramid networks for object detection.
\newblock In {\em CVPR}.

\bibitem[\protect\citeauthoryear{Lin \bgroup et al\mbox.\egroup
  }{2017b}]{lin2017focal}
Lin, T.-Y.; Goyal, P.; Girshick, R.; He, K.; and Doll{\'a}r, P.
\newblock 2017b.
\newblock Focal loss for dense object detection.
\newblock In {\em ICCV}.

\bibitem[\protect\citeauthoryear{Liu \bgroup et al\mbox.\egroup
  }{2019}]{liu2019auto}
Liu, C.; Chen, L.-C.; Schroff, F.; Adam, H.; Hua, W.; Yuille, A.; and Fei-Fei,
  L.
\newblock 2019.
\newblock Auto-deeplab: Hierarchical neural architecture search for semantic
  image segmentation.
\newblock In {\em CVPR}.

\bibitem[\protect\citeauthoryear{Liu, Huang, and
  others}{2018}]{liu2018receptive}
Liu, S.; Huang, D.; et~al.
\newblock 2018.
\newblock Receptive field block net for accurate and fast object detection.
\newblock In {\em ECCV}.

\bibitem[\protect\citeauthoryear{Liu, Simonyan, and Yang}{2019}]{liu2018darts}
Liu, H.; Simonyan, K.; and Yang, Y.
\newblock 2019.
\newblock Darts: Differentiable architecture search.
\newblock In {\em ICLR}.

\bibitem[\protect\citeauthoryear{Miikkulainen \bgroup et al\mbox.\egroup
  }{2019}]{miikkulainen2019evolving}
Miikkulainen, R.; Liang, J.; Meyerson, E.; Rawal, A.; Fink, D.; Francon, O.;
  Raju, B.; Shahrzad, H.; Navruzyan, A.; Duffy, N.; et~al.
\newblock 2019.
\newblock Evolving deep neural networks.
\newblock In {\em Artificial Intelligence in the Age of Neural Networks and
  Brain Computing}. Elsevier.

\bibitem[\protect\citeauthoryear{Nekrasov \bgroup et al\mbox.\egroup
  }{2019}]{nekrasov2018fast}
Nekrasov, V.; Chen, H.; Shen, C.; and Reid, I.
\newblock 2019.
\newblock Fast neural architecture search of compact semantic segmentation
  models via auxiliary cells.
\newblock In {\em CVPR}.

\bibitem[\protect\citeauthoryear{Paszke \bgroup et al\mbox.\egroup
  }{2017}]{paszke2017automatic}
Paszke, A.; Gross, S.; Chintala, S.; Chanan, G.; Yang, E.; DeVito, Z.; Lin, Z.;
  Desmaison, A.; Antiga, L.; and Lerer, A.
\newblock 2017.
\newblock Automatic differentiation in pytorch.
\newblock In {\em NerIPS Workshop}.

\bibitem[\protect\citeauthoryear{Ren \bgroup et al\mbox.\egroup
  }{2015}]{ren2015faster}
Ren, S.; He, K.; Girshick, R.; and Sun, J.
\newblock 2015.
\newblock Faster r-cnn: Towards real-time object detection with region proposal
  networks.
\newblock In {\em NerIPS}.

\bibitem[\protect\citeauthoryear{Ronneberger, Fischer, and
  Brox}{2015}]{ronneberger2015u}
Ronneberger, O.; Fischer, P.; and Brox, T.
\newblock 2015.
\newblock U-net: Convolutional networks for biomedical image segmentation.
\newblock In {\em MICCAI}.

\bibitem[\protect\citeauthoryear{Roth \bgroup et al\mbox.\egroup
  }{2015}]{roth2015improving}
Roth, H.~R.; Lu, L.; Liu, J.; Yao, J.; Seff, A.; Cherry, K.; Kim, L.; and
  Summers, R.~M.
\newblock 2015.
\newblock Improving computer-aided detection using convolutional neural
  networks and random view aggregation.
\newblock {\em IEEE transactions on medical imaging} 35(5):1170--1181.

\bibitem[\protect\citeauthoryear{Russakovsky \bgroup et al\mbox.\egroup
  }{2015}]{russakovsky2015imagenet}
Russakovsky, O.; Deng, J.; Su, H.; Krause, J.; Satheesh, S.; Ma, S.; Huang, Z.;
  Karpathy, A.; Khosla, A.; Bernstein, M.; et~al.
\newblock 2015.
\newblock Imagenet large scale visual recognition challenge.
\newblock {\em International Journal of Computer Vision} 115(3):211--252.

\bibitem[\protect\citeauthoryear{Salakhutdinov, Torralba, and
  Tenenbaum}{2011}]{salakhutdinov2011learning}
Salakhutdinov, R.; Torralba, A.; and Tenenbaum, J.
\newblock 2011.
\newblock Learning to share visual appearance for multiclass object detection.
\newblock In {\em CVPR}.

\bibitem[\protect\citeauthoryear{Shin \bgroup et al\mbox.\egroup
  }{2016}]{shin2016deep}
Shin, H.-C.; Roth, H.~R.; Gao, M.; Lu, L.; Xu, Z.; Nogues, I.; Yao, J.;
  Mollura, D.; and Summers, R.~M.
\newblock 2016.
\newblock Deep convolutional neural networks for computer-aided detection: Cnn
  architectures, dataset characteristics and transfer learning.
\newblock {\em IEEE transactions on medical imaging} 35(5):1285--1298.

\bibitem[\protect\citeauthoryear{Szegedy \bgroup et al\mbox.\egroup
  }{2016}]{szegedy2016rethinking}
Szegedy, C.; Vanhoucke, V.; Ioffe, S.; Shlens, J.; and Wojna, Z.
\newblock 2016.
\newblock Rethinking the inception architecture for computer vision.
\newblock In {\em CVPR}.

\bibitem[\protect\citeauthoryear{Tajbakhsh \bgroup et al\mbox.\egroup
  }{2016}]{tajbakhsh2016convolutional}
Tajbakhsh, N.; Shin, J.~Y.; Gurudu, S.~R.; Hurst, R.~T.; Kendall, C.~B.;
  Gotway, M.~B.; and Liang, J.
\newblock 2016.
\newblock Convolutional neural networks for medical image analysis: Full
  training or fine tuning?
\newblock {\em IEEE transactions on medical imaging} 35(5):1299--1312.

\bibitem[\protect\citeauthoryear{Wang \bgroup et al\mbox.\egroup
  }{2017}]{wang2017zoom}
Wang, Z.; Yin, Y.; Shi, J.; Fang, W.; Li, H.; and Wang, X.
\newblock 2017.
\newblock Zoom-in-net: Deep mining lesions for diabetic retinopathy detection.
\newblock In {\em MICCAI}.

\bibitem[\protect\citeauthoryear{Yan, Bagheri, and Summers}{2018}]{yan20183d}
Yan, K.; Bagheri, M.; and Summers, R.~M.
\newblock 2018.
\newblock 3d context enhanced region-based convolutional neural network for
  end-to-end lesion detection.
\newblock In {\em MICCAI}.

\bibitem[\protect\citeauthoryear{Yan \bgroup et al\mbox.\egroup
  }{2018a}]{yan2018deeplesion}
Yan, K.; Wang, X.; Lu, L.; and Summers, R.~M.
\newblock 2018a.
\newblock Deeplesion: automated mining of large-scale lesion annotations and
  universal lesion detection with deep learning.
\newblock {\em Journal of Medical Imaging} 5(3):036501.

\bibitem[\protect\citeauthoryear{Yan \bgroup et al\mbox.\egroup
  }{2018b}]{yan2018deep}
Yan, K.; Wang, X.; Lu, L.; Zhang, L.; Harrison, A.~P.; Bagheri, M.; and
  Summers, R.~M.
\newblock 2018b.
\newblock Deep lesion graphs in the wild: relationship learning and
  organization of significant radiology image findings in a diverse large-scale
  lesion database.
\newblock In {\em CVPR}.

\bibitem[\protect\citeauthoryear{Yu \bgroup et al\mbox.\egroup
  }{2018}]{yu2018recurrent}
Yu, Q.; Xie, L.; Wang, Y.; Zhou, Y.; Fishman, E.~K.; and Yuille, A.~L.
\newblock 2018.
\newblock Recurrent saliency transformation network: Incorporating multi-stage
  visual cues for small organ segmentation.
\newblock In {\em CVPR}.

\bibitem[\protect\citeauthoryear{Zhu \bgroup et al\mbox.\egroup
  }{2019}]{zhu2018deformable}
Zhu, X.; Hu, H.; Lin, S.; and Dai, J.
\newblock 2019.
\newblock Deformable convnets v2: More deformable, better results.
\newblock In {\em CVPR}.

\bibitem[\protect\citeauthoryear{Zoph and Le}{2017}]{zoph2016neural}
Zoph, B., and Le, Q.~V.
\newblock 2017.
\newblock Neural architecture search with reinforcement learning.
\newblock In {\em ICLR}.

\bibitem[\protect\citeauthoryear{Zoph \bgroup et al\mbox.\egroup
  }{2018}]{zoph2018learning}
Zoph, B.; Vasudevan, V.; Shlens, J.; and Le, Q.~V.
\newblock 2018.
\newblock Learning transferable architectures for scalable image recognition.
\newblock In {\em CVPR}.

\end{thebibliography}
\end{document}